\documentclass[conference]{IEEEtran}
\IEEEoverridecommandlockouts
\usepackage{cite}
\usepackage{url}
\usepackage{amsmath,amssymb,amsfonts}
\usepackage{algorithmic}
\usepackage{subfigure}
\usepackage{multirow}
\usepackage{graphicx}
\usepackage{textcomp}
\usepackage{booktabs}
\usepackage{flushend}
\usepackage{xcolor}
\def\BibTeX{{\rm B\kern-.05em{\sc i\kern-.025em b}\kern-.08em
    T\kern-.1667em\lower.7ex\hbox{E}\kern-.125emX}}
\begin{document}

\def\eg{\textit{e.g}. } \def\Eg{\textit{E.g}. }
\def\ie{\textit{i.e}. } \def\Ie{\textit{I.e}. }
\def\cf{\textit{c.f}. } \def\Cf{\textit{C.f}. }
\def\etc{\textit{etc}. } \def\vs{\textit{vs}. }
\def\etcn{\textit{etc}.}
\def\wrt{w.r.t. } \def\etal{\textit{et al}. }

\title{Crowd--powered Face Manipulation Detection: Fusing Human Examiner Decisions}

\author{\IEEEauthorblockN{C. Rathgeb, R. Nichols, M. Ibsen, P. Drozdowski,  C. Busch}
\IEEEauthorblockA{\textit{da/sec -- Biometrics and Internet Security Research Group} \\
Hochschule Darmstadt, Germany\\
\small{\texttt{\{christian.rathgeb,robert.nichols,mathias.ibsen,christoph.busch\}@h-da.de}}}
}

\maketitle

%
\begin{abstract}
We investigate the potential of fusing human examiner decisions for the task of digital face manipulation detection. To this end, various decision fusion methods are proposed incorporating the examiners' decision confidence, experience level, and their time to take a decision. Conducted experiments are based on a psychophysical evaluation of digital face image manipulation detection capabilities of humans in which different manipulation techniques were applied, \ie face morphing, face swapping and retouching. The decisions of 223 participants were fused to simulate crowds of up to seven human examiners. Experimental results reveal that (1) despite the moderate detection performance achieved by single human examiners, a high accuracy can be obtained through decision fusion and (2) a weighted fusion which takes the examiners' decision confidence into account yields the most competitive detection performance.
\end{abstract}


\section{Introduction}
Digital face manipulation \cite{2021_Book_DigitalFaceManipulation} has rapidly advanced in the recent past and numerous facial alteration methods have been proposed, \eg face swapping or morphing. Digitally manipulated face images can be misused for malicious purposes, \eg document fraud or spreading of misinformation. Hence, harms caused by face image manipulation may lead to a loss of trust in digital content. In response to this, different research laboratories proposed various automated face manipulation detection systems, for surveys on this topic refer to \cite{9115874,TOLOSANA2020131}. Public face manipulation detection benchmarks have already been established, \eg in \cite{roessler2019faceforensics,jiang2020deeperforensics1}, and encouraging results have been reported. However, many state-of-the-art detection methods have been revealed to lack generalisability and hence obtain only moderate performance in challenging real-world scenarios, \eg in the presence of image post-processing or unseen manipulations. 

\begin{figure}[!t]
\vspace{-0.0cm}
\centering
\includegraphics[width=\linewidth]{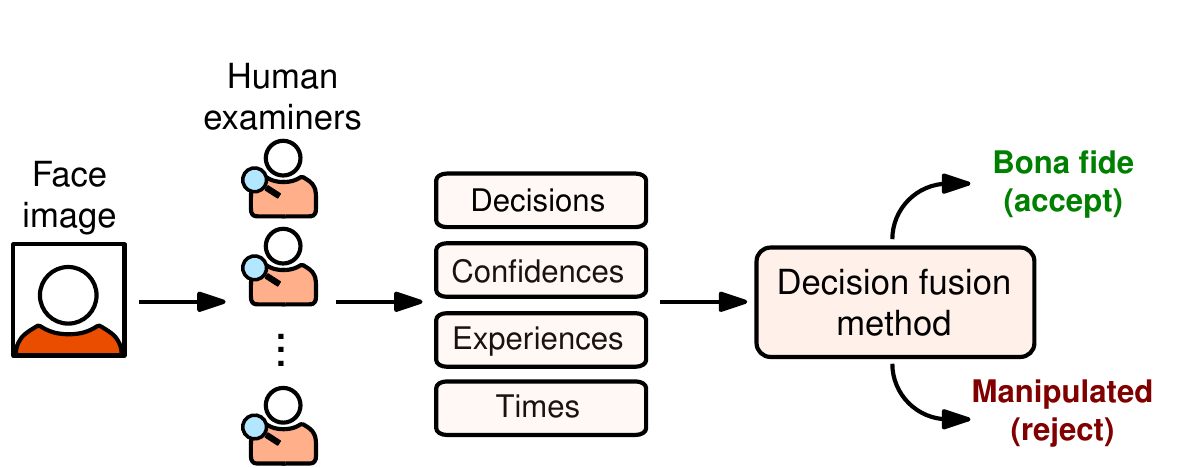}\label{fig:original1}
\caption{Overview of the crowd-powered face manipulation detection: for a the task of detecting a manipulated face image, the detection decisions, corresponding confidence scores, experience and required time of several human examiners are fused to obtain a final decision.}\label{fig:example}\vspace{-0.2cm}
\end{figure}

In a biometric scenario, such as automated border control, a human inspector compares a presented facial image against a trusted live capture. Studies  have consistently shown individual performance in face recognition to be widely distributed across the population \cite{burton:2010,dowsett:2015}. Given practical and theoretical implications, determining the source of these differences has become a relevant issue, in turn prompting development of standardised tests such as the Glasgow Face Matching Test (GMFT \cite{burton:2010}) and the Cambridge Face Perception Test (CFPT \cite{Duchaine:2006}), amongst others. 

Similar to automated detection systems, human performance for detecting digitally altered face images has been analysed in some previous works, \eg in \cite{ferrara:2016,makrushin:2017,Robertson2017,kramer:2019,korschunov:2020, robertson:2018,nightingale:2021,zhang:2021,Grohe2110013119,nichols2022psychophysical}. While research in the area of human face manipulation detection is relatively sparse, initial studies have indicated that humans display inadequate proficiency in detecting certain types of manipulated face images. Related works usually focus on one type of manipulation, \eg morphing, in addition to unquantified differences in quality of manipulations, which greatly complicates comparison and reduces practical relevance. Moreover, some works use incomplete or erroneous implementation of test procedures, in some cases modifying well studied procedures, resulting in unknown effects on robustness of findings. Furthermore, related studies do not consider an elaborated fusion of human detection results.

Towards achieving more reliable face image manipulation detection, the potential of fusing human examiner decisions is investigated in this work. For this purpose, experiments have been conducted based on a psychophysical evaluation of digital face image manipulation detection capabilities considering different popular face manipulation techniques. Obtained decisions of individual human examiners are then combined to groups of various sizes and fused considering additional parameters, \ie confidence, experience, and time, see Fig.~\ref{fig:example}. To this end, various decision fusion approaches are considered, including majority voting as well as confidence-weighted, experience-weighted, and time-weighted fusion. It is shown that fusing the decisions of human examiners generally yields significantly improved detection performance. In particular, a decision fusion weighted by the confidence of human examiners' achieves the most competitive results. 

This paper is organised as follows:  methods employed to fuse decisions are described in section~\ref{sec:fusion}. The experimental setup is summarised in section~\ref{sec:setup} and obtained results are presented in section~\ref{sec:results}. Finally, conclusions are drawn in section~\ref{sec:conclusion}.

\section{Fusion Methods}\label{sec:fusion}
We assume that an odd number of $k$ humans examine a potentially manipulated face image. Each examiner has to decide whether a face image is manipulated or not, \ie bona fide. This results in a binary decision vector $\mathbf{d}=(d_i)_{i=1}^k$, $d_i \in \{0,1\}$, where 0 may indicate \emph{bona fide} and 1 \emph{manipulated}, respectively. In addition, it is assumed that each examiner provides a decision confidence score resulting in confidence vector $\mathbf{c}=(c_i)_{i=1}^k$ with $c_{i} \in \{1\dots C\}$, where higher values reflect higher confidence. Similarly, each examiner indicates an experience level resulting in an experience vector $\mathbf{e}=(e_i)_{i=1}^k$ with $e_{i} \in \{1\dots E\}$, where higher values reflect greater experience. Finally, let $\mathbf{t}=(t_i)_{i=1}^k$ with $t_i \in \mathbb{N}$ denote the vector of times required by the examiners to take their decision. 

Based on the aforementioned inputs we consider different fusion methods which are detailed in the following subsections.

\subsection{Majority Voting}
In this approach, the choice (decision) that receives the majority of the vote is the winner. Majority voting can improve the overall detection performance by eliminating less frequent erroneous decisions. In the binary classification of face image manipulation, a majority vote ($\mathit{MV}$) can be easily estimated based on the decision vector solely, 
\begin{equation}
 \mathit{MV}(\mathbf{d})=\left\{
\begin{array}{ll}
0 & \textrm{if } \sum_{i=1}^k d_{i} < k/2\textrm{,}\\
1 & \, \textrm{otherwise.} \\
\end{array}
\right. 
\end{equation}

\subsection{Confidence-weighted Fusion}
This method extends the majority voting by assigning more weight to examiners who are more confident in their decision. That is, the confidence-weighted fusion can potentially improve the detection performance if decisions taken with higher confidence have generally a higher probability of being correct. The confidence-weighted fusion ($\mathit{CF}$) is calculated as,
\begin{equation}
\mathit{CF}(\mathbf{d},\mathbf{c})=\left\{
\begin{array}{ll}
0 & \textrm{if } \sum_{i=1}^k c_{i} d_{i}  < \sum_{i=1}^k c_{i}(1-d_{i})\textrm{,} \\
1 & \, \textrm{otherwise.} \\
\end{array}
\right. 
\end{equation}

\subsection{Experience-weighted Fusion}
The experience-weighted fusion ($\mathit{EF}$) is the same as $\mathit{CF}$, but uses the experience vector $\mathbf{e}$ as input instead of the confidence vector $\mathbf{c}$. Hence, this method puts more weight on those examiners who are more experienced. Performance is expected to improve, provided that examiners with higher experience achieve higher detection performance. Note that $\mathbf{e}$ consists of static values that are independent of the presented image.

\begin{figure}[!t]
\vspace{-0.0cm}
\centering
\subfigure[Sex]{\includegraphics[width=0.965\linewidth]{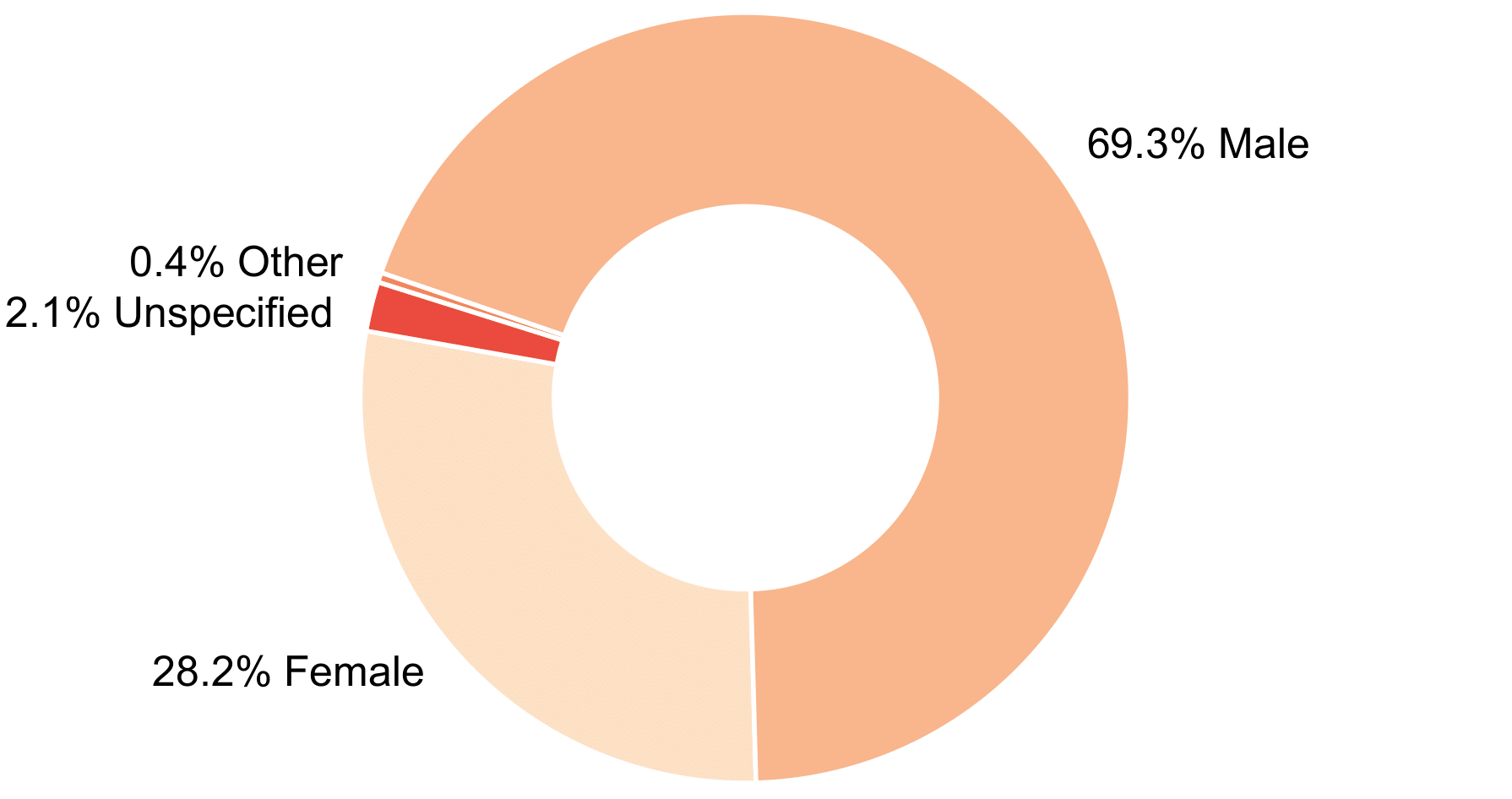}}
\subfigure[Age]{\includegraphics[width=0.95\linewidth]{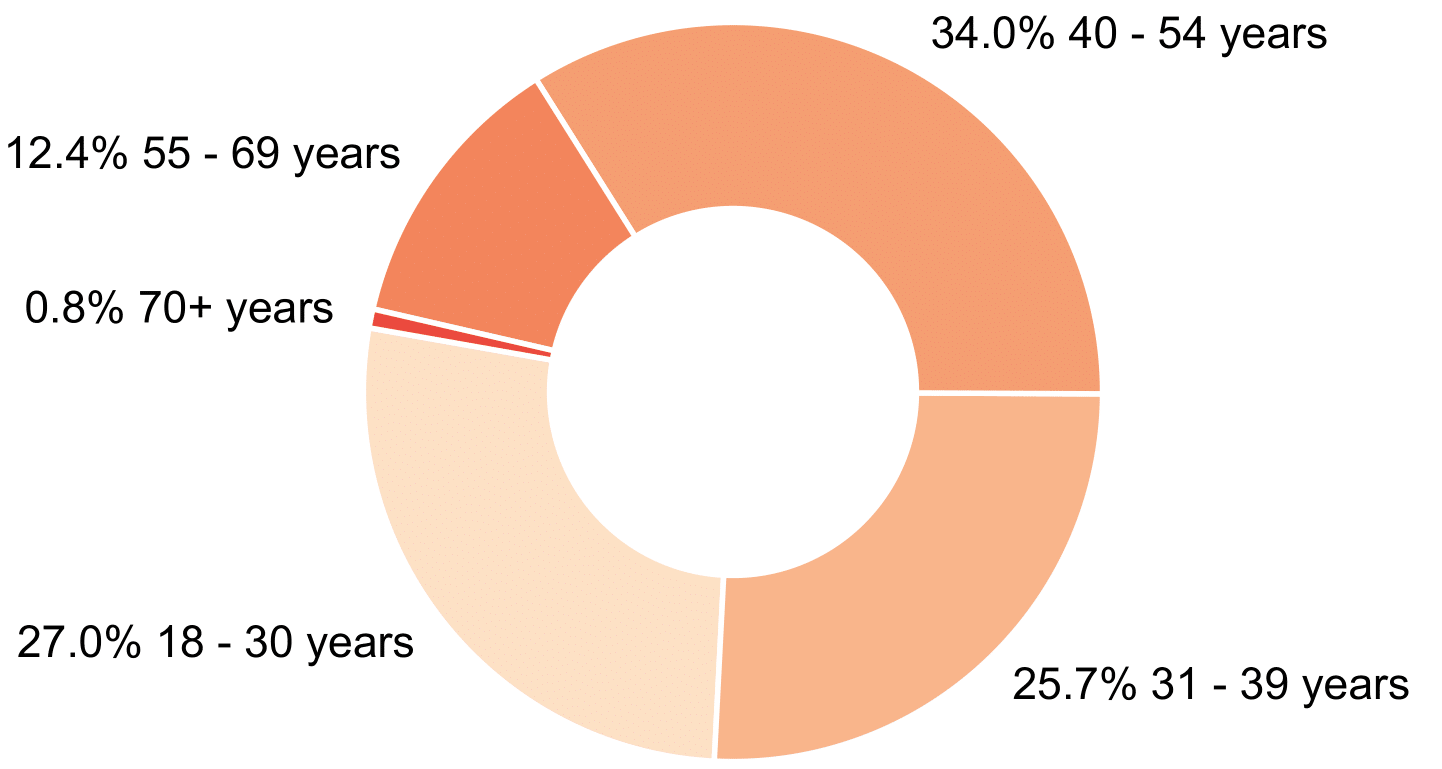}}
\caption{Demographic distribution of participants. }\label{fig:participants}\vspace{-0.2cm}
\end{figure}

\subsection{Time-based Fusion}
This method favours decisions that are made with more time. If, on average, slower decisions have a higher probability of being correct, the overall detection performance will improve. The time-based fusion ($\mathit{TF}$) is similar to $\mathit{CF}$, but uses the time vector $\mathbf{t}$ as input (in seconds). 

\subsection{Overall Fusion}
The aforementioned decision criteria likely exhibit strong correlation. For instance, highly experienced examiners may take decisions faster and with higher confidence. Nevertheless, an overall fusion based on all of the aforementioned criteria may further improve the detection performance. As, confidence, experience, and time of decisions are usually in different ranges and of different types, a direct fusion of these values would require normalisation (based on a training stage). In order to avoid this, majority voting can again be applied to obtain an overall fusion ($\mathit{OF}$) as,
\begin{equation}
\mathit{OF}(\mathbf{d},\mathbf{c},\mathbf{e},\mathbf{t})=\mathit{MV}\big(\mathit{CF}(\mathbf{d},\mathbf{c}), \mathit{EF}(\mathbf{d},\mathbf{e}), \mathit{TF}(\mathbf{d},\mathbf{t})\big)\textrm{.}
\end{equation}
In other words, we apply a majority vote to different fusion-based decisions, thus, no normalisation is required. 


\section{Experimental Setup}\label{sec:setup}

This section briefly describes the experimental setup. Conducted experiments builds upon the work of \cite{nichols2022psychophysical} to which the interested reader is referred to for more details. In order to fuse decisions of human examiners, they first participated in an experiment for detecting digital face image manipulations. To this end, face image manipulations were applied to images from public face image databases \cite{Phillips1998,Phillips2005}. Finally, blocs (crowds) of human examiners were simulated by grouping the participants' results.



\begin{table}[!t]
\centering
\caption{Values assigned to experience and confidence levels.}\label{tab:confidence_experience}\vspace{-0.1cm}
\resizebox{0.8\columnwidth}{!}{
\begin{tabular}{ccc}
\toprule
\textbf{Value} & \textbf{Experience} & \textbf{Confidence} \\
\midrule
1 & none & very unsure \\
2 & basic & unsure \\
3 & intermediate & neutral \\
4 & expert & sure \\
5 & specialised professional & very sure \\
\bottomrule
\end{tabular}
}
\end{table}
\subsection{Human Examiner Face Manipulation Detection}

In an online application called \emph{unclassifyd}\footnote{\url{https://dasec.h-da.de/unclassifyd}}, 223 participants completed 50 experiment trials with the task of detecting digital face image manipulations. In the beginning, participants were asked to provide demographic information about them and to indicate their experience level with digital face image manipulation. Distributions of demographic properties of the participants are depicted in Fig.~\ref{fig:participants}. 

Two types of trials were considered:
\begin{itemize}
\item A-B-X discrimination task (ABX): participants are \emph{sequentially} presented with two known samples (A and B), \ie a bona fide and a manipulated face image, followed by one unknown sample (X).
\item Spatial 2-alternative forced-choice task (S2AFC): participants are \emph{simultaneously} presented with two unknown samples, \ie a bona fide and a manipulated face image.
\end{itemize}

The trials were divided among both types, 23 ABX and 27 S2AFC. Each trial consisted of three components with a textual introduction with a generous timeout. Face images were displayed for 8 seconds side by side in S2AFC trials, and for 6 seconds in sequential arrangement in ABX trials respectively, and at last a vote component to request participant classification and indication of confidence level with a time limit of 60 seconds. Values assigned to the different possible experience levels and confidence scores are listed in Tab.~\ref{tab:confidence_experience}. Feedback was displayed to the participants, both after each trial and upon completion of the entire experiment.

\subsection{Face Image Selection and Manipulation}

\begin{figure}[!t]
\vspace{-0.0cm}
\centering
\subfigure[Morphing]{\includegraphics[height=3cm]{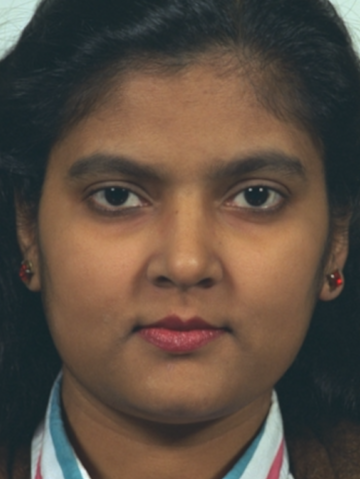} \includegraphics[height=3cm]{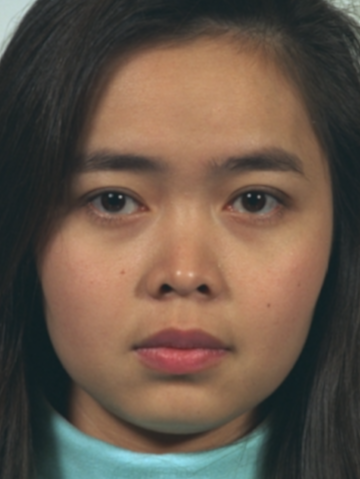} \includegraphics[height=3cm]{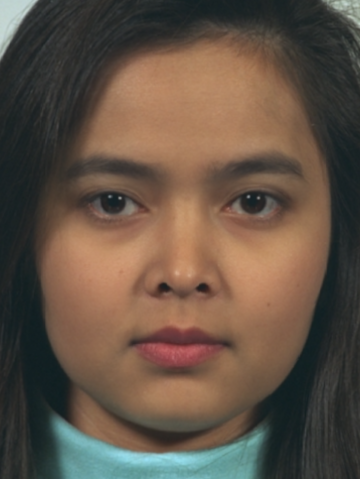}}
\subfigure[Swapping]{\includegraphics[height=3cm]{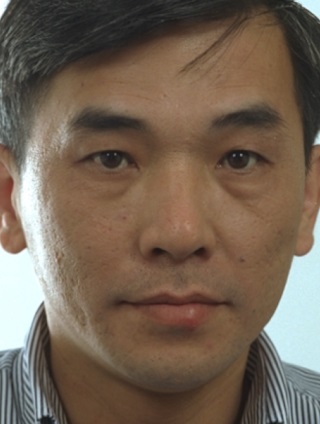} \includegraphics[height=3cm]{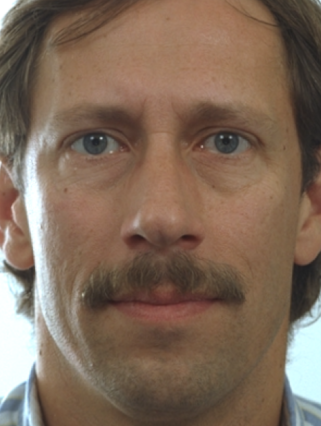} \includegraphics[height=3cm]{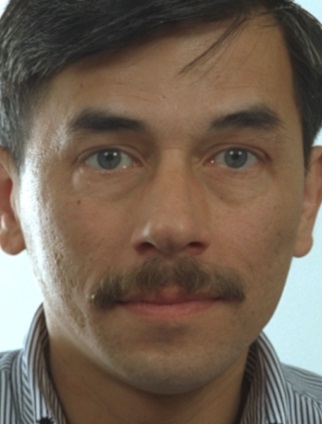}}
\subfigure[Retouching]{\includegraphics[height=3cm]{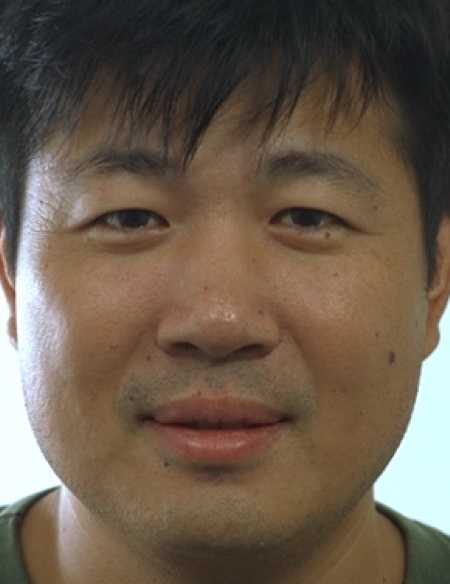} \includegraphics[height=3cm]{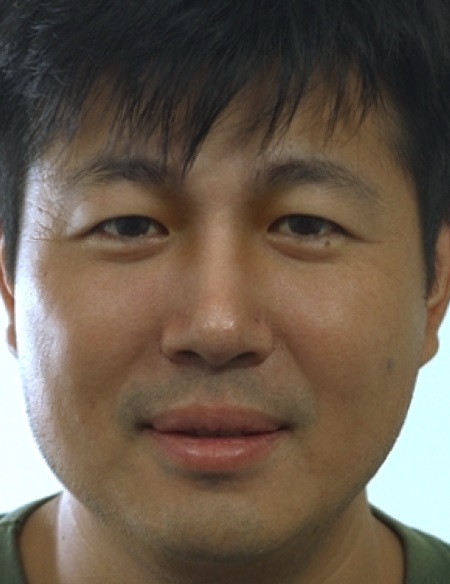}}
\caption{Examples of bona fide and manipulated (rightmost) face images. }\label{fig:examples}\vspace{-0.2cm}
\end{figure}

Bona fide facial images were taken from both the FERET \cite{Phillips1998} and FRGCv2 \cite{Phillips2005}
face databases. The utilised database of manipulated images was created by applying three types of face image manipulations, \ie morphing, swapping, and retouching, each with two different methods resulting in six distinct manipulation classes. Morphing mainly consists of warping and alpha-blending and combines the facial features of two subjects to create an artificial intermediate face image. In contrast, swapping completely exchanges the facial features of one subject with those of another subject. The morphing and swapping process retain the outer face region of one of the subjects. The goal of retouching is to beautify the face of a subject, \eg by thinning the nose or enlarging the eyes. Example images are shown in Fig.~\ref{fig:examples}. These six classes are further divided into easy and hard categories, by calculated distance scores resulting from a comparison with the corresponding bona fide source image. To this end, a pre-trained model of the well-known  ArcFace facial recognition system \cite{Deng19ab} was used (LResNet100E-IR,ArcFace@ms1m-refine-v2). A systematic selection of ten images from each class and difficulty level was performed, resulting in a database of 60 manipulated face images. Subsequently, 75 bona fide face images were automatically selected. Based on this face image set, trials were generated.

\subsection{Grouping Human Examiners}
Crowds of human examiners, \ie teams, were simulated by grouping results $k=$ 3, 5, and 7 participants. For each value of $k$, 1,000 groups were created by random selection (with replacement). That is, of the total number of 223 participants, many were part of several groups.

\begin{figure}[!t]
\vspace{-0.0cm}
\centering
\includegraphics[width=8.75cm]{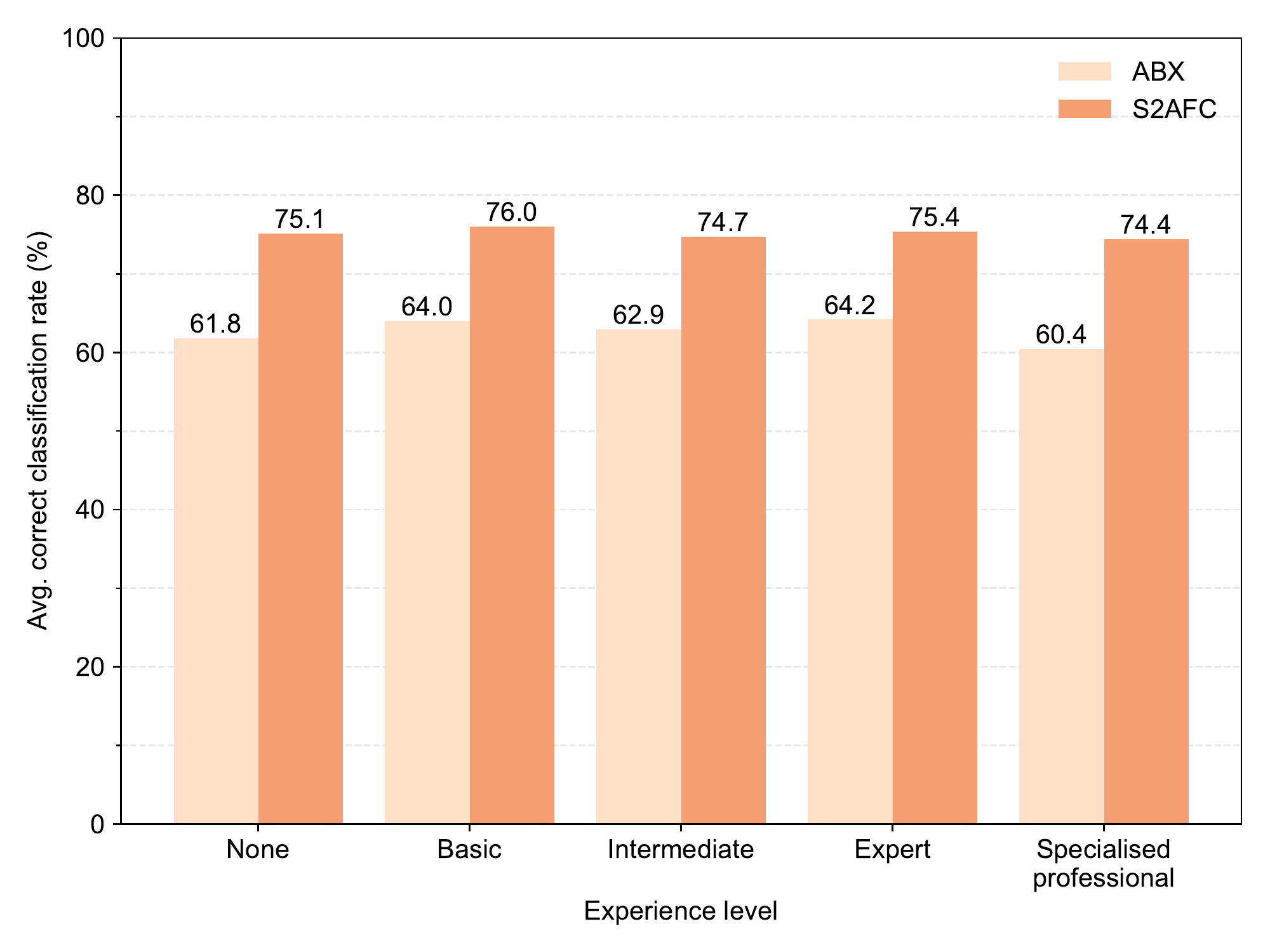}\vspace{-0.2cm}
\caption{Detection performance in relation to experience. }\label{fig:experience}\vspace{-0.2cm}
\end{figure}


\begin{figure}[!t]
\vspace{-0.2cm}
\centering
\subfigure[Confidence]{\includegraphics[height=6.85cm]{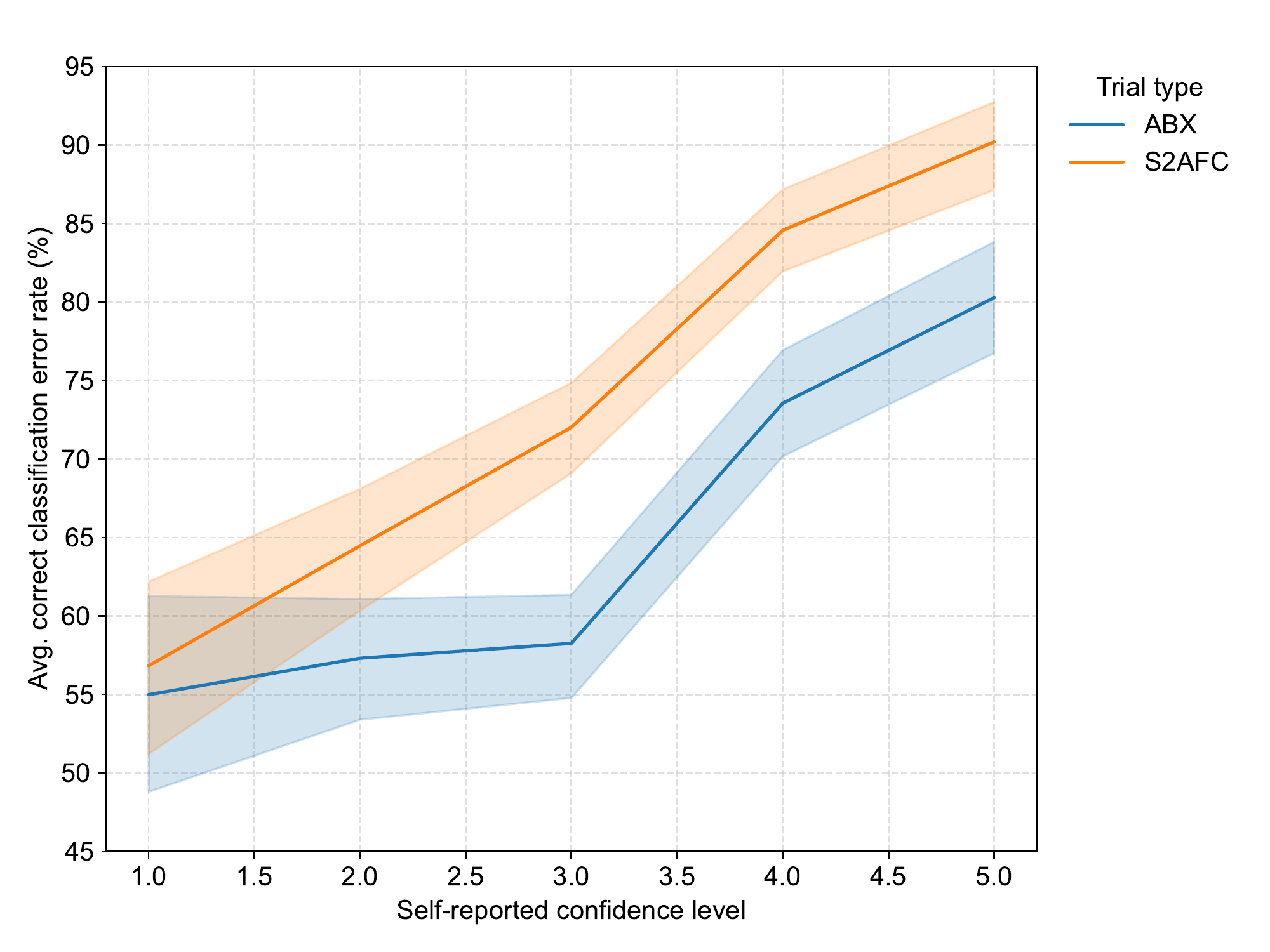}\label{fig:conf}}\hspace{-0.3cm}
\subfigure[Time]{\includegraphics[height=6.85cm]{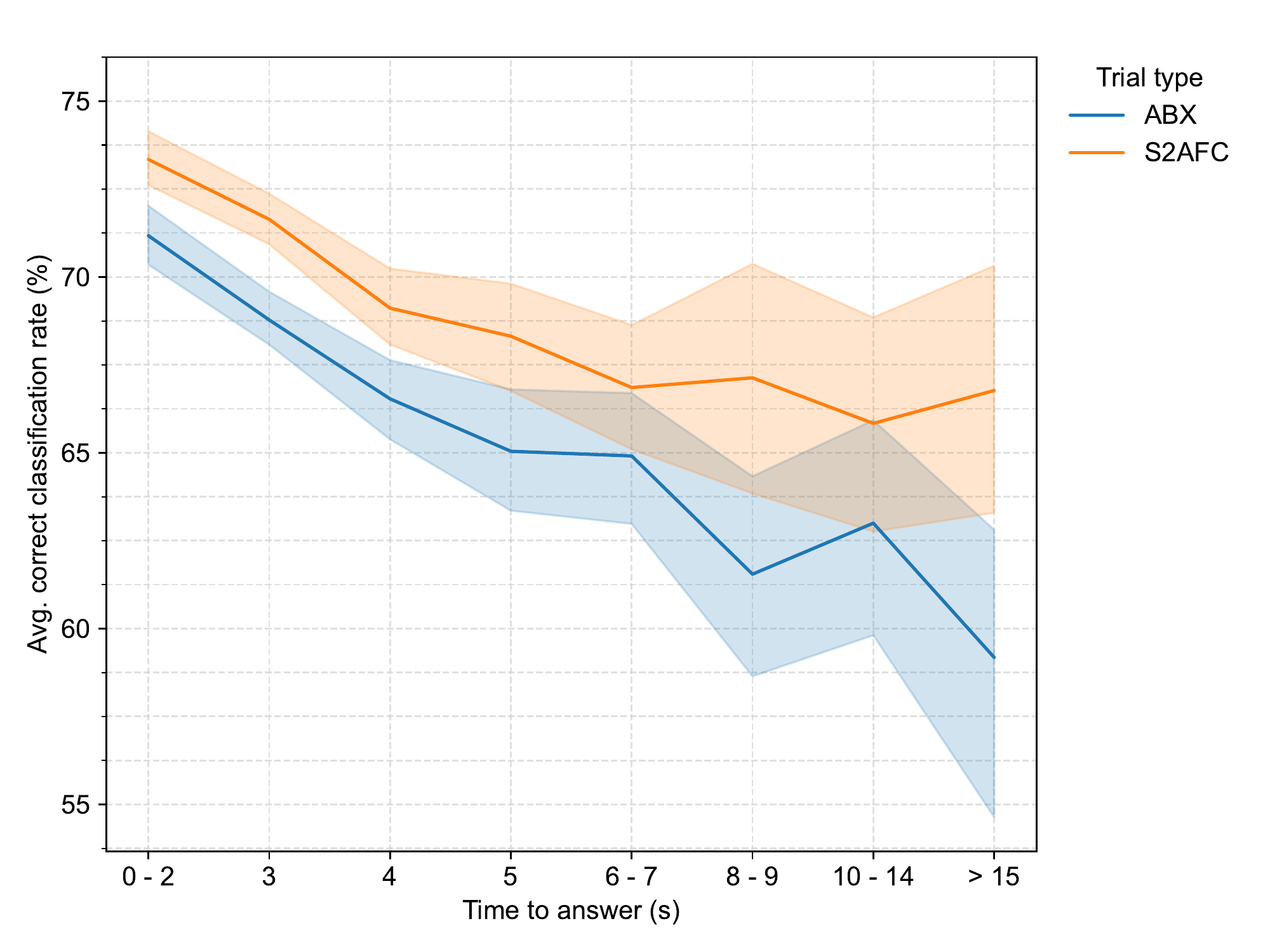}\label{fig:time}}\vspace{-0.2cm}
\caption{Detection performance in relation to confidence and time. }\label{fig:conf_time}\vspace{-0.4cm}
\end{figure}

\begin{table*}
\centering
\small
\caption{Performance of different fusion methods in terms of average CCR and standard deviation (in \%). }\label{tab:results}\vspace{-0.2cm}
\begin{tabular}{lcccccc}
\toprule
\multirow{2}{*}{\textbf{Fusion}}  & \multicolumn{2}{c}{$k=3$} & \multicolumn{2}{c}{$k=5$}  & \multicolumn{2}{c}{$k=7$}  \\ \cmidrule(r){2-3} \cmidrule(r){4-5} \cmidrule{6-7}
  &  \textbf{ABX} &  \textbf{S2AFC} &  \textbf{ABX} &  \textbf{S2AFC} &  \textbf{ABX} &  \textbf{S2AFC}\\\midrule
$\mathit{MV}$ & 67.9 ($\pm$9.5) & 81.8 ($\pm$8.3) & 70.5 ($\pm$9.1) & 85.1 ($\pm$7.2) & 72.2 ($\pm$8.6) & 86.9 ($\pm$5.9) \\
$\mathit{CF}$ & \textbf{68.2 ($\pm$9.3)} & \textbf{82.3 ($\pm$8.3)} & \textbf{71.1 ($\pm$8.9)} & \textbf{85.9 ($\pm$7.1)} & \textbf{73.0 ($\pm$8.3)} & \textbf{87.8 ($\pm$5.6)} \\
$\mathit{EF}$ & 66.2 ($\pm$10.6) & 79.4 ($\pm$10.4) & 68.8 ($\pm$9.5) & 83.1 ($\pm$8.0) & 70.9 ($\pm$9.2) & 85.4 ($\pm$6.7)\\
$\mathit{TF}$ & 65.9 ($\pm$9.9) & 75.2 ($\pm$10.9) & 67.0 ($\pm$9.5) & 81.8 ($\pm$7.6) & 68.7 ($\pm$9.1) & 83.4 ($\pm$6.5)\\
$\mathit{OF}$ & 67.7 ($\pm$9.6) & 81.8 ($\pm$8.1) & 70.1 ($\pm$9.0) & 85.1 ($\pm$7.2) & 71.8 ($\pm$8.8) & 86.9 ($\pm$6.0) \\\bottomrule
\end{tabular}\vspace{-0.5cm}
\end{table*}

\begin{figure*}[!t]
\vspace{-0.0cm}
\centering
\subfigure[$\mathit{MV}$ -- ABX]{\includegraphics[height=5.25cm]{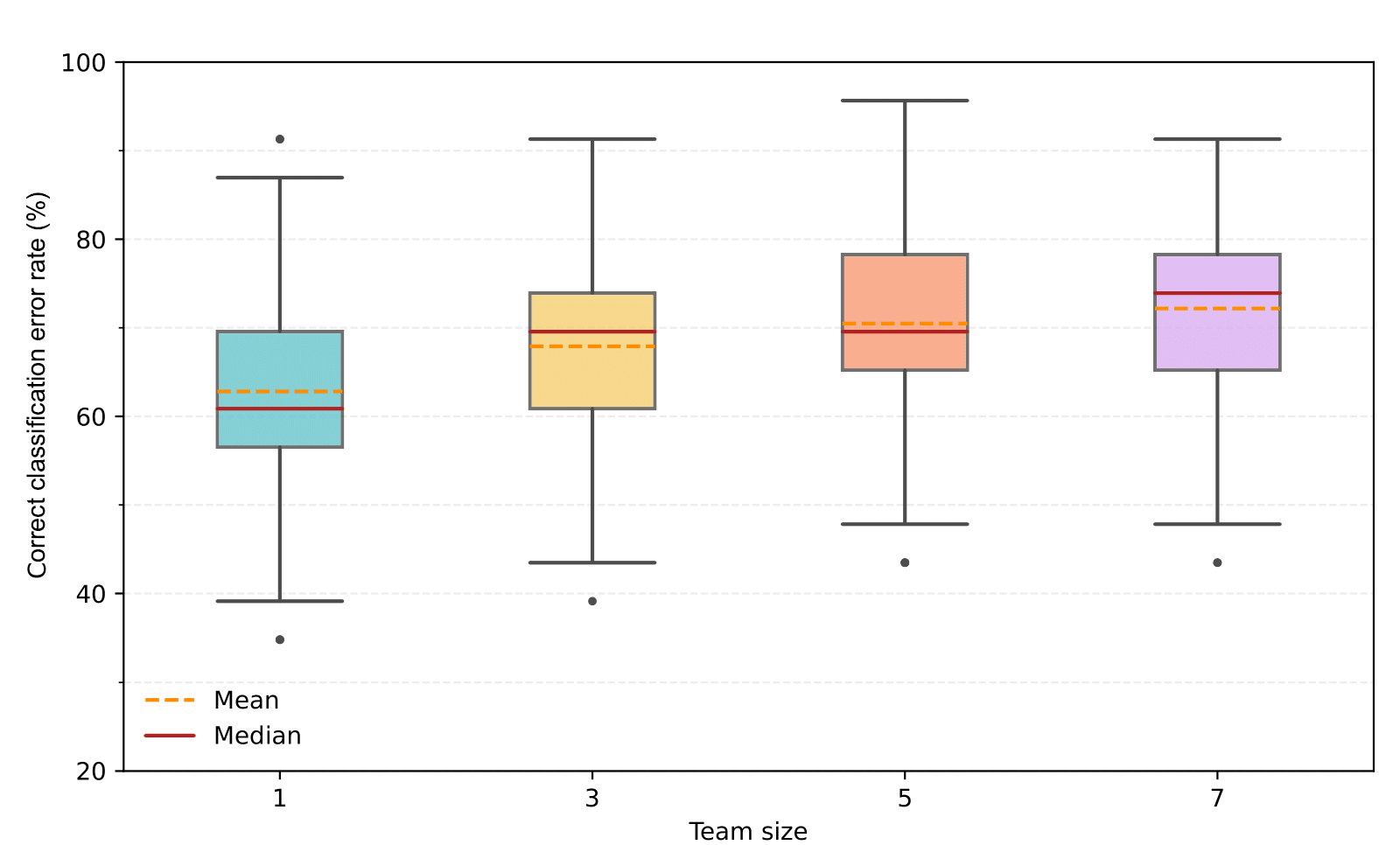}}
\subfigure[$\mathit{MV}$ -- S2AFC]{\includegraphics[height=5.25cm]{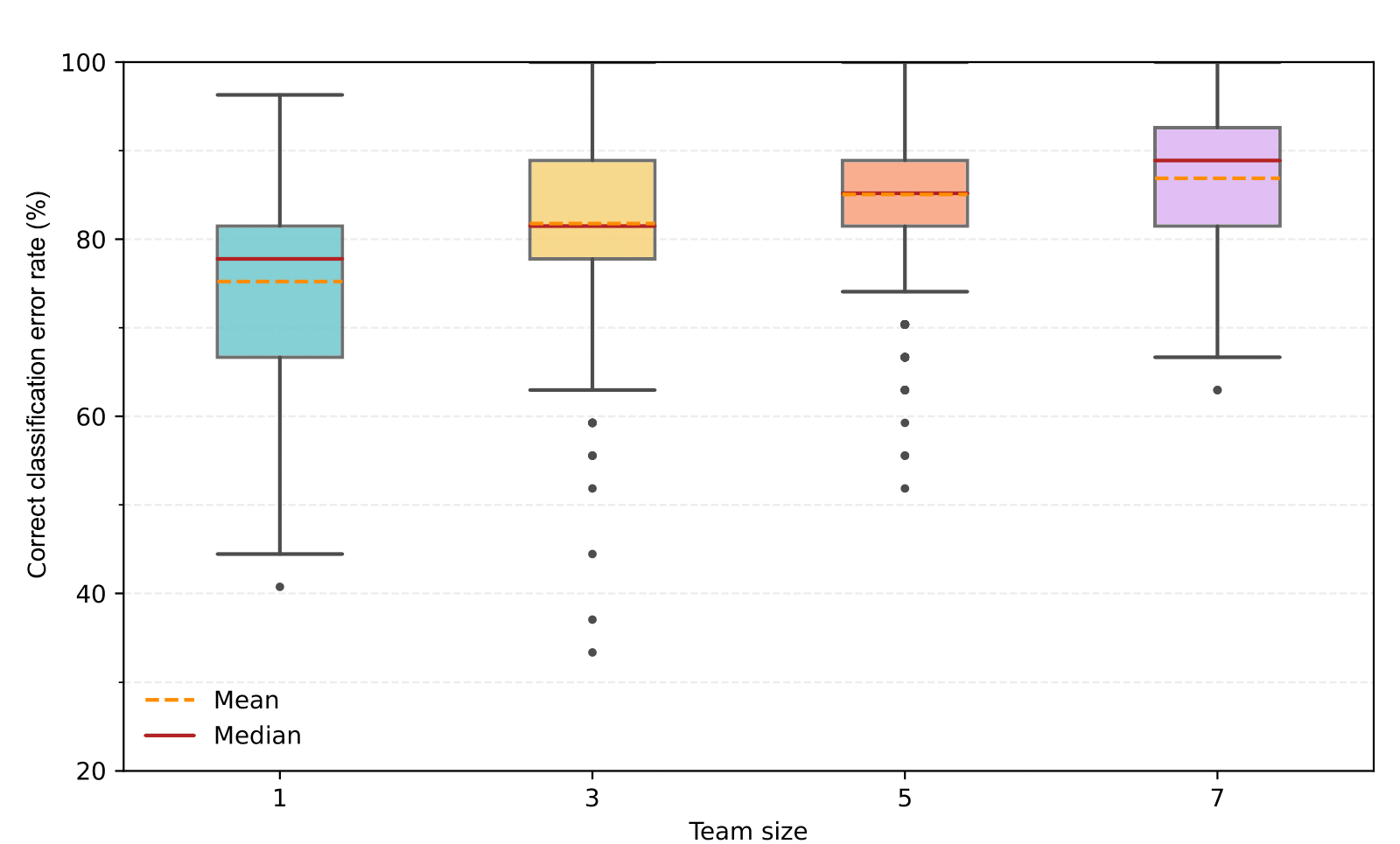}}
\subfigure[$\mathit{CF}$ -- ABX]{\includegraphics[height=5.25cm]{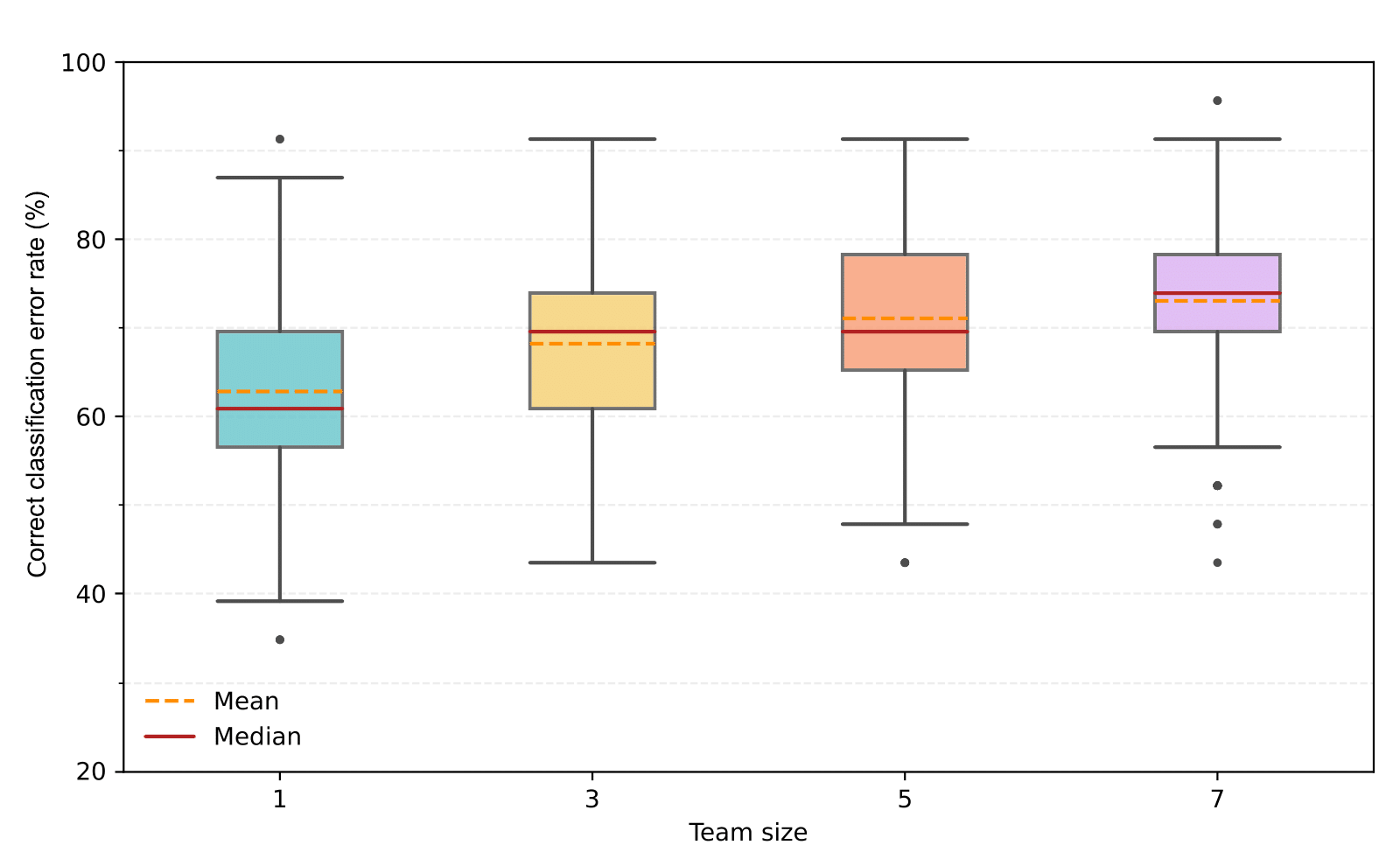}}
\subfigure[$\mathit{CF}$ -- S2AFC]{\includegraphics[height=5.25cm]{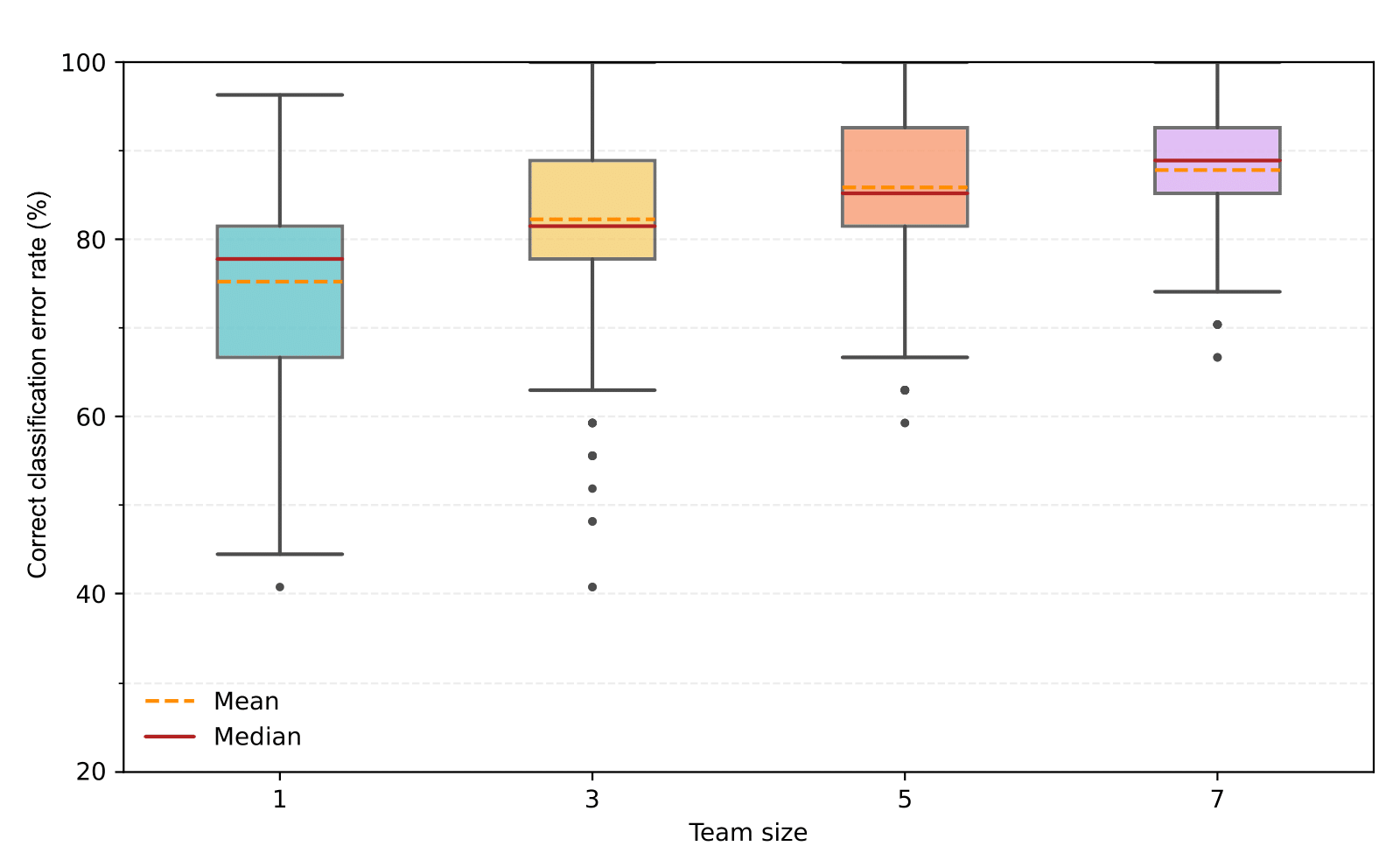}}
\caption{Comparison of majority voting and confidence-based fusion. }\label{fig:cftf}\vspace{-0.4cm}
\end{figure*}

\section{Results}\label{sec:results}
Firstly, the detection performance of the 223 single human examiners was evaluated in terms of correct classification rate (CCR). A moderate performance of 62.8\% ($\pm$11.0) is obtained in ABX trials. Higher detection accuracy of 75.2\% ($\pm$10.9) is achieved for the S2AFC trials. In order to investigate the potential of the proposed fusion methods, the correlation of detection accuracy with confidence, experience, and time is estimated. The detection performance across different levels of experience is depicted in Fig.~\ref{fig:experience}. Interestingly, the detection accuracy does not correlate with the experience levels that the participants had to indicate before the start of the detection task. Note that indicated experience does not relate to dedicated training for the detection task which might further improve the detection accuracy. 

Average detection performance along with its standard deviation in relation to confidence and time is shown in Fig.~\ref{fig:conf_time}. A strong correlation between the confidence level and the correctness of a decision is observable (\cf Fig.~\ref{fig:conf}). Further, a strong negative correlation between the time to take a decision and its correctness seems to be observable where the deviation of detection accuracy significantly increases with time (\cf Fig.~\ref{fig:time}). At first sight, fast decisions generally reveal higher detection accuracy. On the one hand, fast decisions are usually made for manipulations which are easy to detect. Slow decisions may be taken for manipulations which are hard to detect. In the latter case, however, the large variations regarding the correctness of decisions may hamper an effective fusion. It is worth noting that the detection capabilities significantly varied depending on the type of face image manipulation. For instance, face swapping was generally detected with higher accuracy compared to face morphing or retouching. Moreover, strong variations \wrt error types, \ie false positives and false negatives, could be observed for the different face image manipulation types.

Secondly, the performance rates obtained by different fusion methods are summarised in Tab.~\ref{tab:results} (best results are marked bold). Again, higher detection accuracies are generally achieved for the S2AFC trials. High performance gains are observable for the simple fusion based on majority voting ($\mathit{MV}$). For groups of $k=3$ human examiners, detection performance is improved by more than 5 percentage points compared to accuracy obtained by individual human examiners. Further improvements of up to approximately 10 percent points are achieved for groups of $k=5$ and $7$. Additionally, the standard deviation is significantly reduced for larger groups, \ie fused decisions become more robust. 

Across all considered groups sizes and both trial types, the highest detection performance is obtained for the confidence-based fusion method ($\mathit{CF}$). That is, incorporating the decision confidence into the fusion procedure can be beneficial. However, in our experiments, $\mathit{CF}$ only marginally outperforms $\mathit{MV}$ with improvements in the range of 0.5 to 1 percent points. Fig.~\ref{fig:cftf} compares the box plots of detection performance accuracies for the $\mathit{MV}$ and $\mathit{CF}$ method across group sizes for both trial types. It can be seen that the performance rates are very similar while the $\mathit{CF}$ method tends to reduce some outliers. 

As expected, the experience-based fusion method ($\mathit{EF}$) does not obtain competitive results. Precisely, the detection performance rates slightly drop (1 to 2 percentage points) compared to the simple $\mathit{MV}$ method. Similarly, the time-based fusion method ($\mathit{TF}$) reveals clearly inferior detection performance rates.  
 
Lastly, the overall fusion ($\mathit{OF}$) achieves performance rates similar to the $\mathit{MV}$ approach. In this method, the increase and decrease in detection performance resulting from the $\mathit{CF}$ and  $\mathit{TF}$ method, respectively, seem to cancel each other. 


\section{Conclusion}\label{sec:conclusion}
Forensic image analysis tasks, \eg digital face manipulation detection, may often be conducted by a group of human examiners. In such cases, their decisions can be combined to obtain a reliable and robust final decision. However, such a decision fusion may not be trivial since many additional parameters may need to be considered. 

In this work, we showed that human face manipulation detection performance can be greatly improved by fusing the decisions of a group of examiners. In such a scenario,  a weighted fusion which takes  the  examiners’ decision confidence into account was found to reveal the most competitive detection  performance. 


\section*{Acknowledgments}
This work has in part received funding from the European Union’s Horizon 2020  research and innovation programme under the Marie Skłodowska-Curie grant agreement No. 860813 TReSPAsS-ETN and the German Federal Ministry of Education and Research and the Hessian Ministry of Higher Education, Research, Science and the Arts within their joint support of the National Research Center for Applied Cybersecurity ATHENE.

\bibliographystyle{IEEEtran}
\bibliography{references}

\end{document}